\documentclass[12pt]{article}
\usepackage[margin=1in]{geometry}
\usepackage[slantedGreek]{mathpazo}
\usepackage[pdftex]{graphicx}
\usepackage{amsfonts}
\usepackage{amssymb}
\usepackage{amsthm}
\usepackage{graphicx,float}
\usepackage{amsmath}%
\usepackage{hyperref}
\usepackage{natbib}
\usepackage{color}
\usepackage{algorithm}
\usepackage[noend]{algpseudocode}
\title{Generative Bayesian Hyperparameter Tuning}
\author{	
	\makebox[.4\linewidth]{Hedibert Lopes}\\\textit{Department of Statistics}\\\textit{INSPER}\\\and 
	\makebox[.4\linewidth]{Nick Polson}\\\textit{ Chicago Booth}\\\textit{  University of Chicago}\\\and 
	\makebox[.4\linewidth]{Vadim Sokolov\footnote{Hedibert Lopes is Professor of Statistics and Econometrics at INSPER. Nick Polson is Professor of Econometrics and Statistics at Chicago Booth, University of Chicago: ngp@chicagobooth.edu. Vadim Sokolov is Associate Professor at Volgenau School of Engineering, George Mason University, USA: vsokolov@gmu.edu.}}\\\textit{ Department of Systems Engineering }\\\textit{  and Operations Research}\\\textit{ George Mason University}
}

\date{First Version: 20 January 2025\\This Version: \today}

\graphicspath{{./fig/}}

\begin{document}

\maketitle
\begin{abstract}
\noindent Hyper-parameter selection is a central practical problem in modern machine learning, governing regularization strength, model capacity, and robustness choices. Cross-validation is often computationally prohibitive at scale, while fully Bayesian hyper-parameter learning can be difficult due to the cost of posterior sampling. We develop a generative perspective on hyper-parameter tuning that combines two ideas: (i) optimization-based approximations to Bayesian posteriors via randomized, weighted objectives (weighted Bayesian bootstrap), and (ii) amortization of repeated optimization across many hyper-parameter settings by learning a transport map from hyper-parameters (including random weights) to the corresponding optimizer. This yields a “generator look-up table’’ for estimators, enabling rapid evaluation over grids or continuous ranges of hyper-parameters and supporting both predictive tuning objectives and approximate Bayesian uncertainty quantification. We connect this viewpoint to weighted $M$-estimation, envelope/auxiliary-variable representations that reduce non-quadratic losses to weighted least squares, and recent generative samplers for weighted $M$-estimators.
\end{abstract}

\section{Introduction}

Hyper-parameters such as regularization strengths, robustness/quantile levels, and architectural or optimization settings have an outsized impact on predictive performance and calibration. In practice, these choices are frequently made by cross-validation (CV), despite well-known statistical and computational drawbacks in some regimes \citep{stone1974crossvalidatory}. Bayesian hyper-parameter learning offers a principled alternative, but can be challenging in modern models because posterior computation is expensive and must typically be repeated for each hyper-parameter configuration.

This paper develops a generative viewpoint on hyper-parameter tuning \citep{nareklishvili2024generative,polson2025generative,nareklishvili2023generative}. We start from optimization formulations of regularized estimation and their MAP interpretations, and then leverage randomized, weighted objectives to obtain scalable approximate posteriors via the weighted Bayesian bootstrap (WBB) \citep{newton2021weighted}. The key additional step is amortization: rather than re-solving the optimization problem from scratch for each hyper-parameter value, we learn a transport map (generator) from hyper-parameters and random weights to the corresponding optimizer. This idea is closely related in spirit to hyper-networks and amortized inference, where one learns to predict parameters or approximate posterior objects as a function of context \citep{ha2017hypernetworks,bhadra2024merging}.

Viewed as a hyper-parameter optimization strategy, the learned map can be used to rapidly evaluate predictive criteria over hyper-parameter grids (a drop-in replacement for expensive CV loops) or to support gradient-based outer-loop tuning, complementing black-box Bayesian optimization \citep{snoek2012practical,bergstra2012random,baker2022analyzing} and gradient-based/bilevel approaches \citep{maclaurin2015gradient,franceschi2018bilevel}. Viewed as approximate Bayesian inference, combining WBB with a generator yields fast sampling of parameter draws across hyper-parameter settings.

The remainder of the paper develops the core ingredients and illustrates how envelope/auxiliary-variable representations can turn difficult losses into weighted least-squares inner steps, aligning them with the same computational template as WBB and the generative map.

\subsection{Existing literature}

Two strands of literature are especially close to our unifying perspective. First, \citet{golub1979generalized} develop generalized cross-validation (GCV) as an analytic, rotation-invariant surrogate for leave-one-out cross-validation in ridge regression and related linear smoother settings. Their key observation is that when fitted values are linear in $y$ through a hat matrix $A(\lambda)$, one can estimate predictive risk using only residuals and the effective degrees of freedom $\mathrm{tr}\{A(\lambda)\}$, yielding a tuning rule that avoids explicit refitting across held-out folds and does not require estimating $\sigma^2$. In modern terms, GCV provides an explicit outer criterion $U(\cdot)$ for $\lambda$ in a setting where the inner optimizer map $\lambda\mapsto \hat{\theta}(\lambda)$ is available in closed form.

Second, \citet{shin2023generative,shin2024generative} propose the Generative Multi-purpose Sampler (GMS), which constructs a generator $G(w,\lambda,\eta)$ that outputs solutions of weighted (penalized) $M$-estimators across bootstrap and cross-validation weight patterns and across tuning parameters. A central contribution is their training objective: instead of learning $G$ by regressing onto precomputed optimizers (which would require many expensive inner solves), they minimize an integrated weighted objective value over draws of $(w,\lambda,\eta)$, so that generator training and $M$-estimation co-evolve via a single optimization loop. They also demonstrate how this enables computationally intensive procedures (iterated bootstrap, bootstrapped CV, solution paths) at scale.

Our contribution is to place these ideas into a single paradigm suitable for modern models trained by SGD: a randomized weighted inner objective, an outer tuning criterion defined as an expectation over the induced randomness, and an optional transport map/generator to amortize repeated optimization. In this view, GCV is the deterministic linear-smoother special case where both the optimizer map and the outer risk surrogate are analytic, while GMS is the weighted $M$-estimation special case where the generator is trained by criterion minimization. The rest of the paper develops this unification in the context of WBB-based approximate posteriors and fast uncertainty summaries.

\paragraph{Contributions and scope.}
This paper contributes a unifying template that connects classical analytic tuning (GCV), weighted $M$-estimation perturbation schemes (bootstrap/CV weights), and generator-based amortization. We provide a concrete algorithm box that makes the mapping $(\omega,\lambda,\eta)\mapsto\hat{\theta}$ operational for SGD-trained models, and we demonstrate its performance in both controlled ridge regression and modern deep-learning settings (MNIST). We show how criterion-based training can avoid expensive optimizer-label generation while still enabling fast evaluation of tuning objectives and uncertainty summaries. Code to reproduce all experiments is available at \url{https://github.com/VadimSokolov/Generative-Bayesian-Hyperparameter-Tuning}.

\section{Generative Bayesian Hyperparameter Tuning}

\subsection{A unified method that generalizes GCV and GMS}

We now state a generic method that subsumes (i) classical analytic tuning rules such as GCV for linear smoothers \citep{golub1979generalized} and (ii) modern generator-based weighted $M$-estimation as in GMS \citep{shin2023generative,shin2024generative}. The method has three components: an \emph{inner} (possibly randomized) training objective, an \emph{outer} tuning criterion, and an optional \emph{transport map} (generator) that amortizes repeated solutions of the inner problem.

\paragraph{Inner problem (randomized weighted training objective).}
Let $h=(\lambda,\eta)$ denote a vector of hyper-parameters: $\lambda$ is a regularization strength and $\eta$ collects any additional hyper-parameters that index the data-fit term (e.g.\ robustness/quantile parameters, temperature parameters, augmentation strengths, etc.). Let $\omega\sim\pi(\omega)$ be random weights (including deterministic $\omega\equiv \mathbf{1}$ as a special case). Define the weighted objective
\begin{equation}
\label{eq:inner}
L(\theta;\omega,h)
\;=\;
\frac{1}{n}\sum_{i=1}^n \omega_i\,\ell_{\eta}(y_i\mid \theta)
\;+\;
\lambda\,\phi(\theta),
\qquad
\hat{\theta}(\omega,h)\in \arg\min_{\theta} L(\theta;\omega,h).
\end{equation}
In neural networks, $\hat{\theta}(\omega,h)$ is computed approximately by SGD on reweighted mini-batches. In convex problems, $\hat{\theta}(\omega,h)$ may be available in closed form.

\paragraph{Transport map / generator (amortizing the optimizer).}
We introduce a parametric map $g_{\varphi}$ intended to approximate the optimizer:
\[
g_{\varphi}(\omega,h)\approx \hat{\theta}(\omega,h).
\]
This can be trained either (A) by supervised regression to precomputed optimizers $\hat{\theta}(\omega^{(b)},h^{(b)})$, or (B) by direct minimization of the criterion value (GMS-style) by plugging $g_{\varphi}$ into $L(\cdot;\omega,h)$ and optimizing $\varphi$ by backpropagation.

\paragraph{Outer problem (what ``hyper-parameter tuning'' means).}
The outer goal is to choose $h$ to optimize a criterion $U$ that measures out-of-sample quality and/or uncertainty. A general template is
\begin{equation}
\label{eq:outer}
\min_{h}\; J(h)
\;=\;
\mathbb{E}_{\omega\sim\pi(\omega)}\Big[U\big(\hat{\theta}(\omega,h),h\big)\Big]
\;\approx\;
\frac{1}{M}\sum_{m=1}^M U\big(g_{\hat{\varphi}}(\omega^{(m)},h),h\big),
\end{equation}
where $U$ might be a validation negative log-likelihood, a proper scoring rule, a risk estimate, or a proxy for marginal-likelihood objectives. The approximation on the right is the computational payoff: once $g_{\hat{\varphi}}$ is trained, we can evaluate (and often differentiate) $J(h)$ without repeatedly re-running SGD for each $h$.

\paragraph{Special case 1 (Golub--Heath--Wahba GCV).}
In ridge regression (or any linear smoother), the optimizer is explicit and \eqref{eq:inner} is deterministic: take $\omega\equiv \mathbf{1}$ and $\eta$ empty, with squared-error loss and quadratic penalty. Then $\hat{\theta}(h)$ is available in closed form and the fitted values are linear in $y$ via a hat matrix $A(\lambda)$. Choosing $U$ in \eqref{eq:outer} as the generalized cross-validation surrogate yields
\[
\hat{\lambda}\in \arg\min_{\lambda} V(\lambda),
\]
which is precisely the classical GCV rule \citep{golub1979generalized}. In this case, one may view $g(\omega,h)$ as the exact analytic map $h\mapsto\hat{\theta}(h)$ (no learned generator needed), and $U$ as an analytic plug-in estimate of predictive risk.

\paragraph{Special case 2 (Shin--Wang--Liu GMS).}
In weighted $M$-estimation, \eqref{eq:inner} matches the GMS setting: $\omega$ is random (bootstrap/Dirichlet-type weights), $\ell_{\eta}$ is a loss indexed by $\eta$, and $\lambda$ tunes regularization. GMS emphasizes the criterion-based training route: rather than generating labels $\hat{\theta}(\omega^{(b)},h^{(b)})$, one trains $g_{\varphi}$ by minimizing the weighted objective value directly (our mode (B) above). In our unified view, GMS corresponds to a particular choice of $\pi(\omega)$, model class $g_{\varphi}$, and training objective for the transport map; the outer criterion \eqref{eq:outer} can then be used to tune $h$ for prediction (e.g.\ validation risk) or for approximate Bayesian sampling over $\hat{\theta}(\omega,h)$.

\subsection{Illustration with a simple example}
\label{sec:toy-illustration}

This section gives a concrete toy illustration, closely following the setup and motivation in \citet{shin2023generative}. The point is not to advocate linear regression, but to make explicit why training a generator by minimizing an integrated criterion can be more efficient than matching precomputed optimizers.

\paragraph{Weighted ridge regression as a map $(\omega,\lambda)\mapsto\hat{\theta}$.}
Consider a linear model with fixed design matrix $X\in\mathbb{R}^{n\times p}$ and response $y\in\mathbb{R}^n$. Let $\omega\in\mathbb{R}^n_+$ be weights and $W=\mathrm{diag}(\omega)$. The weighted ridge objective is
\[
L(\theta;\omega,\lambda)
\;=\;
\frac{1}{n}\|W^{1/2}(y-X\theta)\|_2^2 + \lambda\|\theta\|_2^2,
\]
with closed-form optimizer
\[
\hat{\theta}(\omega,\lambda)
\;=\;
\left(X^\top W X + n\lambda I\right)^{-1}X^\top W y.
\]
Thus, in this toy setting the ``true'' map $(\omega,\lambda)\mapsto\hat{\theta}(\omega,\lambda)$ exists explicitly, and bootstrap/CV-type procedures correspond to repeatedly evaluating this map for many draws of $\omega$ and many values of $\lambda$.

\paragraph{Two ways to train a generator.}
Let $g_{\varphi}(\omega,\lambda)$ be a neural network intended to approximate $\hat{\theta}(\omega,\lambda)$. A classical supervised approach would generate a training set $\{(\omega^{(b)},\lambda^{(b)},\hat{\theta}^{(b)})\}_{b=1}^B$ by repeatedly computing $\hat{\theta}^{(b)}=\hat{\theta}(\omega^{(b)},\lambda^{(b)})$, then fit $\varphi$ by regression (e.g.\ squared loss). This is accurate when $B$ is very large, but becomes expensive when computing $\hat{\theta}^{(b)}$ is costly (e.g.\ neural networks).

The GMS approach instead trains the generator by minimizing the integrated objective value, without ever computing $\hat{\theta}(\omega,\lambda)$ labels:
\begin{equation}
\label{eq:gms-toy}
\min_{\varphi}\; \mathbb{E}_{(\omega,\lambda)\sim P}\Big[L\big(g_{\varphi}(\omega,\lambda);\omega,\lambda\big)\Big],
\end{equation}
where $P$ is a user-chosen distribution over weights and tuning parameters. In practice, the expectation is approximated by Monte Carlo samples inside SGD: on each iteration one draws fresh $(\omega,\lambda)$ and backpropagates through $L(g_{\varphi}(\omega,\lambda);\omega,\lambda)$.

\paragraph{Why criterion-based training can be more sample-efficient.}
Supervised training controls $\|g_{\varphi}(\omega,\lambda)-\hat{\theta}(\omega,\lambda)\|$ only on the sampled training set, and can overfit if $B$ is limited. By contrast, the criterion-based loss \eqref{eq:gms-toy} supplies a task-relevant signal at essentially unlimited ``training inputs'' $(\omega,\lambda)$ because drawing new $(\omega,\lambda)$ is cheap; the optimization of $\varphi$ and the minimization of the statistical objective co-evolve in a single loop \citep{shin2023generative}.

One way to formalize this intuition is to evaluate a population mismatch measure such as an integrated prediction loss,
\[
\mathrm{IPL}(g_{\varphi})
\;=\;
\mathbb{E}_{(\omega,\lambda)\sim P}\Big[\|g_{\varphi}(\omega,\lambda)-\hat{\theta}(\omega,\lambda)\|_2^2\Big],
\]
which measures accuracy of the generator averaged over the weight/tuning distribution of interest. Even in settings where $L$ is convex and has an explicit optimizer, optimizing \eqref{eq:gms-toy} can reduce $\mathrm{IPL}(g_{\varphi})$ efficiently because it provides gradients everywhere in $(\omega,\lambda)$-space, whereas supervised fitting relies on a finite (and expensive) set of optimizer labels.

\begin{figure}[t]
\centering
\includegraphics[width=0.85\linewidth]{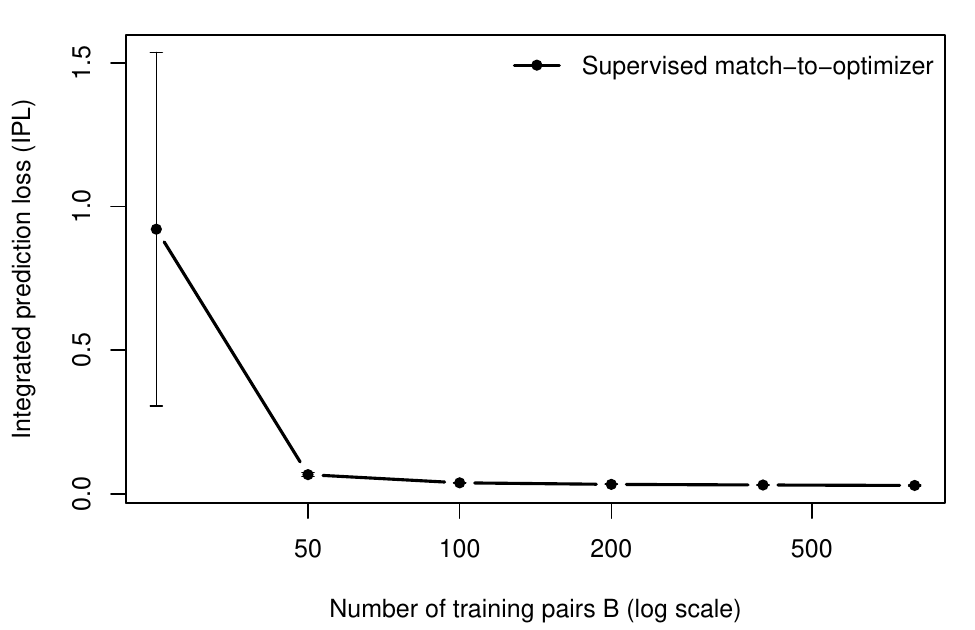}
\caption{Toy weighted-ridge example. Integrated prediction loss (IPL) as a function of the number of labeled optimizer solves $B$ used by supervised match-to-optimizer training. Criterion-based (GMS-style) training uses unlabeled Monte Carlo draws and is essentially flat in $B$.}
\label{fig:toy-gms-ipl}
\end{figure}

\paragraph{Toy computation (R).}
To make this contrast concrete, we implement the weighted ridge setup in R (script: \texttt{code/toy/toy\_gms\_ridge.R}). We compare two linear-generator training modes: (A) supervised fitting that uses $B$ optimizer labels $\hat{\theta}(\omega^{(b)},\lambda^{(b)})$ and (B) criterion-based fitting that uses many Monte Carlo draws $(\omega,\lambda)$ without computing any optimizer labels (mimicking the GMS advantage in problems where $\hat{\theta}$ requires an expensive solve).
Figure~\ref{fig:toy-gms-ipl} reports $\mathrm{IPL}$ versus $B$; Table~\ref{tab:toy-gms-ipl} gives representative values.

\begin{table}[t]
\centering
\caption{Toy weighted-ridge example: integrated prediction loss (IPL) for a linear generator trained either by supervised matching to optimizer labels or by criterion-based (GMS-style) training. Values are means over replications with Monte Carlo standard errors in parentheses.}
\label{tab:toy-gms-ipl}
\begin{tabular}{lccc}
\hline
 & \multicolumn{3}{c}{Training pairs $B$}\\
Method & 50 & 200 & 800\\
\hline
Supervised match-to-optimizer & 0.067 (0.003) & 0.033 (0.001) & 0.029 (0.001) \\
Criterion-based (GMS-style) & 0.033 (0.000) & 0.032 (0.000) & 0.031 (0.000) \\
\hline
\end{tabular}
\end{table}

\subsection{Motivation and setup}

Cross validation (CV) is computationally expensive for training. It also has statistical shortcomings in some regimes, including inconsistency \citep{stone1974crossvalidatory}. A Bayesian approach to hyper-parameter tuning is conceptually straightforward but computationally difficult (often requiring MCMC). A central link is the duality between regularization and maximum a posteriori (MAP) inference: we treat $\lambda$ as a hyper-parameter with prior density $\lambda \sim p(\lambda)$.

Many problems in machine learning involve an output $y_i$ with $x_i$ a high-dimensional input. A generic regularized objective takes the form, with $\theta=(\theta_1,\ldots,\theta_p)$,
\[
\ell(\theta;\lambda)=\sum_{i=1}^n \ell\big(y_i, f_{\theta}(x_i)\big) + \lambda \sum_{j=1}^p \phi(\theta_j).
\]
Here $\ell(y_i,f_\theta(x_i))$ denotes a data-fit term (e.g.\ negative log-likelihood or a surrogate loss) and $\phi$ is a regularizer; throughout, we treat the data $\{(x_i,y_i)\}_{i=1}^n$ as fixed and study how the optimizer varies with hyper-parameters.
In the simplest (unpenalized) case, an $M$-estimator can be written as $\hat{\theta}=\arg\min_{\theta}\,\ell(\theta)$, where for likelihood-based problems one often has
\[
\ell(\theta)= -\frac{1}{n}\sum_{i=1}^n \log p(y_i\mid \theta).
\]
Hyper-parameters can enter either through the regularization term (e.g.\ $\lambda$) or through auxiliary/nuisance parameters in the data-fit term (write $\ell_\eta$ for a loss indexed by $\eta$).

\subsection{MAP as an optimization problem}

The MAP estimator can be viewed as the posterior mode for a probabilistic model defined by
\begin{align*}
&p(y \mid \theta)\propto \exp\{-\ell(y, f_{\theta}(x))\},\; \;  p(\theta)\propto \exp\{-\lambda\phi(\theta)\},\\
&p(\theta \mid y)\propto  \exp\{-\ell(y, f_{\theta}(x))-\lambda \phi(\theta)\}.
\end{align*}
Here $p(\theta)$ can be interpreted as a prior distribution and the log-prior as the regularization penalty \citep{datta2025bayesian}.

\subsection{Generator / transport-map viewpoint}

We propose generator methods \citep{polson2025generative,nareklishvili2023deep}: rather than solving the optimization from scratch for each hyper-parameter configuration, learn a map that approximates the optimizer. For weighted objectives (bootstrap-type perturbations), with $\omega=(\omega_1,\ldots,\omega_n)\sim \pi(\omega)$, we can view
\[
G_\phi(\omega,\lambda,\eta) \approx \hat{\theta}(\omega,\lambda,\eta),
\]
as a parametric transport map from hyper-parameters (including random weights) to an approximate optimizer.

This viewpoint suggests a ``generative WBB'' construction: generate training pairs by sampling $(\omega^{(b)},\lambda^{(b)})$ (and potentially $\eta^{(b)}$), computing the corresponding optimizer $\hat{\theta}^{(b)}=\hat{\theta}(\omega^{(b)},\lambda^{(b)},\eta^{(b)})$ via the inner optimization, and then fitting a function $g$ (e.g.\ a neural network) to regress $\hat{\theta}$ on $(\omega,\lambda,\eta)$. A simple surrogate objective is
\[
\hat{g} \in \arg\min_{g}\sum_{b=1}^B \left\|\hat{\theta}^{(b)} - g(\omega^{(b)},\lambda^{(b)})\right\|^2,
\]
which aims to amortize the optimization across draws and hyper-parameter settings.
An important alternative—emphasized by the generative multi-purpose sampler (GMS) for weighted $M$-estimation \citep{shin2023generative,shin2024generative}—is to train the generator by directly minimizing the \emph{criterion value} rather than matching $\hat{\theta}$ labels. In that case one plugs $g_\phi(\omega,\lambda,\eta)$ into the weighted objective and optimizes $\phi$ via backpropagation:
\[
\min_{\phi}\; \frac{1}{B}\sum_{b=1}^B
\left[
\frac{1}{n}\sum_{i=1}^n \omega_i^{(b)}\,\ell_{\eta^{(b)}}\!\big(y_i\mid g_\phi(\omega^{(b)},\lambda^{(b)},\eta^{(b)})\big)
 + \lambda^{(b)} \phi\!\big(g_\phi(\omega^{(b)},\lambda^{(b)},\eta^{(b)})\big)
\right].
\]
This "criterion-based" training can be far cheaper than supervised regression on $\hat{\theta}^{(b)}$, because it avoids repeatedly solving the inner optimization to produce labels; it also aligns naturally with neural networks where $\ell$ is already differentiable and optimized by SGD.
Once such a map is available, hyper-parameter tuning can be posed as an outer optimization problem, for example choosing $\lambda$ (and possibly $\eta$) to minimize an expected predictive loss under the induced randomness,
\[
\min_{\lambda,\eta}\; \mathbb{E}_{\omega \sim \pi(\omega)}\Big[ \mathcal{R}\big(\hat{\theta}(\omega,\lambda,\eta)\big)\Big]
\;\approx\;
\min_{\lambda,\eta}\; \frac{1}{B}\sum_{b=1}^B \mathcal{R}\big(g(\omega^{(b)},\lambda,\eta)\big),
\]
where $\mathcal{R}$ might be a validation loss, a proper scoring rule, or a proxy for marginal-likelihood objectives. This framing highlights why amortization matters: it allows many evaluations (and potentially gradients) with respect to $(\lambda,\eta)$ without repeatedly solving the full inner optimization.
Operationally, one may sample $\lambda^{(b)}$ from a proposal distribution over hyper-parameters (e.g.\ a bounded range, a discrete grid, or a prior $p(\lambda)$), sample weights $\omega^{(b)}\sim\pi(\omega)$, compute $\hat{\theta}^{(b)}$ with any off-the-shelf optimizer (often warm-started across nearby $\lambda$), and then train $g$ on the resulting supervised data. The same template applies when additional hyper-parameters are present, such as a quantile level $q$ in quantile regression.

\paragraph{What we mean by hyper-parameter tuning (neural networks + SGD).}
In modern deep learning, the most direct and practically meaningful notion of hyper-parameter tuning is decision-theoretic: choose $(\lambda,\eta)$ to optimize an out-of-sample predictive criterion \citep{polson2024generative,nareklishvili2023feature}. Concretely, let $\mathcal{D}_{\mathrm{train}}$ denote the training set and $\mathcal{D}_{\mathrm{val}}$ a validation set. For each $(\lambda,\eta)$ and each draw of random weights $\omega\sim\pi(\omega)$, we obtain a trained parameter vector $\hat{\theta}(\omega,\lambda,\eta)$ by (approximately) solving the weighted objective using SGD on mini-batches with per-example reweighting. We then tune $(\lambda,\eta)$ by minimizing an estimated expected validation risk,
\[
J(\lambda,\eta) \;=\; \mathbb{E}_{\omega\sim\pi(\omega)}\Big[\mathcal{R}\big(\hat{\theta}(\omega,\lambda,\eta);\mathcal{D}_{\mathrm{val}}\big)\Big]
\;\approx\; \frac{1}{M}\sum_{m=1}^M \mathcal{R}\big(g(\omega^{(m)},\lambda,\eta);\mathcal{D}_{\mathrm{val}}\big),
\]
where $\mathcal{R}$ is typically a proper scoring rule (e.g.\ negative log-likelihood) to promote both accuracy and calibration, and the last approximation uses the trained generator to avoid re-running SGD many times.

\begin{algorithm}[t]
\caption{Generative WBB for hyper-parameter tuning and uncertainty summaries}
\label{alg:gwbb}
\begin{algorithmic}[1]
\Require Training data $\mathcal{D}_{\mathrm{train}}$, validation data $\mathcal{D}_{\mathrm{val}}$; hyper-parameter proposal $\Pi(\lambda,\eta)$; weight distribution $\pi(\omega)$; samples $B$ (training) and $M$ (evaluation); generator family $g_{\phi}$.
\State \textbf{(Choose generator training mode)} Either (A) supervised regression to match $\hat{\theta}$, or (B) criterion-based training (as in GMS).
\State \textbf{(Training draws)} For $b=1,\ldots,B$:
\State \hspace{1em}Sample $(\lambda^{(b)},\eta^{(b)})\sim \Pi(\lambda,\eta)$ and weights $\omega^{(b)}\sim \pi(\omega)$.
\If{mode (A)}
\State \hspace{1em}Compute $\hat{\theta}^{(b)} \approx \arg\min_{\theta}\Big\{\frac{1}{n}\sum_{i=1}^n \omega_i^{(b)}\,\ell_{\eta^{(b)}}(y_i\mid \theta) + \lambda^{(b)}\phi(\theta)\Big\}$ using SGD on $\mathcal{D}_{\mathrm{train}}$ (reweighted mini-batches).
\State \textbf{(Train generator)} Fit $\hat{\phi}\in\arg\min_{\phi}\sum_{b=1}^B \big\|\hat{\theta}^{(b)} - g_{\phi}(\omega^{(b)},\lambda^{(b)},\eta^{(b)})\big\|^2$.
\Else
\State \textbf{(Train generator)} Fit $\hat{\phi}\in\arg\min_{\phi}\sum_{b=1}^B \Big[\frac{1}{n}\sum_{i=1}^n \omega_i^{(b)}\,\ell_{\eta^{(b)}}(y_i\mid g_{\phi}(\omega^{(b)},\lambda^{(b)},\eta^{(b)})) + \lambda^{(b)}\phi(g_{\phi}(\cdot))\Big]$.
\EndIf
\State \textbf{(Tune hyper-parameters)} For candidate $(\lambda,\eta)$:
\State \hspace{1em}Estimate $J(\lambda,\eta)\approx \frac{1}{M}\sum_{m=1}^M \mathcal{R}(g_{\hat{\phi}}(\omega^{(m)},\lambda,\eta);\mathcal{D}_{\mathrm{val}})$ with $\omega^{(m)}\sim\pi(\omega)$, and set $(\hat{\lambda},\hat{\eta})\in\arg\min_{\lambda,\eta} J(\lambda,\eta)$.
\State \textbf{(Uncertainty summaries)} With tuned $(\hat{\lambda},\hat{\eta})$, sample $\omega^{(m)}\sim\pi(\omega)$, set $\theta^{(m)}=g_{\hat{\phi}}(\omega^{(m)},\hat{\lambda},\hat{\eta})$, and summarize predictive uncertainty using the empirical distribution of $\{p(y\mid x,\theta^{(m)})\}_{m=1}^M$.
\end{algorithmic}
\end{algorithm}

\subsection{Weighted Bayesian bootstrap (WBB)}

Newton et al.\ propose weighted Bayesian bootstrap (WBB) as a way to convert fast scalable optimization into an approximate posterior distribution \citep{newton2021weighted}. They consider random weights (including a penalty weight $\omega_0$) and solve
\[
\hat{\theta} ( \omega , \lambda , Y ).
\]
More generally (and matching weighted $M$-estimation notation), one can write
\[
\hat{\theta}(\omega,\lambda,\eta)
= \arg\min_{\theta}\left\{
\frac{1}{n}\sum_{i=1}^n \omega_i\,\ell_{\eta}(y_i\mid \theta) + \lambda \phi(\theta)
\right\},
\]
which reduces to WBB-type objectives under appropriate choices of $\ell_\eta$ and $\pi(\omega)$ (e.g.\ exponential/Dirichlet/multinomial bootstrap weights).
The weighted log-likelihood in \citep{newton2021weighted} is
\[
\mathcal{L}_\omega(\theta,\lambda,y) = \sum_{i=1}^n \omega_i\, \ell( y_i \mid \theta ) + \lambda \omega_0 \phi(\theta),
\qquad
\phi(\theta) = \sum_{j=1}^p \phi_j(\theta_j),
\]
where $\omega_i=\ln(1/u_i)$ with $u_i\sim U(0,1)$. This induces a map $\omega \mapsto \theta^*_\omega$, and the conjecture is that the induced distribution of $\theta^*_\omega$ (with data fixed) approximates the Bayesian posterior. From a broader perspective, randomized/tempered objectives connect naturally to generalized Bayes and power posterior ideas \citep{bissiri2016general,friel2008power}.

\subsection{Uncertainty summaries}

For neural networks trained by SGD, it is often useful to separate (i) a point estimate used for deployment and (ii) uncertainty summaries used for calibration, risk-sensitive decisions, or model comparison. Under the generative WBB construction, uncertainty arises from the random weights $\omega$: after tuning $(\hat{\lambda},\hat{\eta})$, we draw $\omega^{(m)}\sim\pi(\omega)$, form $\theta^{(m)}=g_{\hat{\phi}}(\omega^{(m)},\hat{\lambda},\hat{\eta})$, and approximate the posterior predictive by Monte Carlo averaging,
\[
p(y\mid x,\mathcal{D}) \;\approx\; \frac{1}{M}\sum_{m=1}^M p(y\mid x,\theta^{(m)}).
\]
This yields simple summaries such as predictive means, variances, and quantiles, and can be combined with standard calibration diagnostics. Conceptually, this plays a role similar to other scalable approximate uncertainty methods for neural networks, such as MC-dropout \citep{gal2016dropout} and deep Gaussian processes \citep{schultz2022deep}, but is aligned with the "random weights + optimization" template used throughout this paper.

\vspace{0.1in}
Bootstrap: $ \omega \sim \mathrm{Multi} ( \iota_n / n ) $

\vspace{0.1in}
Bayesian bootstrap: $ \omega \sim \mathrm{Dirichlet} ( n , \iota_n ) $

\vspace{0.1in}
\noindent A useful related device is the ``quantile trick'' for turning quantile-regression (check) loss into a weighted least-squares form via a variance--mean Gaussian envelope \citep{polson2016mixtures}. For $q\in(0,1)$ and residual $r$, define the (asymmetric) absolute/hinge loss
\[
\rho_q(r) = |r| + (2q-1)r.
\]
\cite{polson2016mixtures} show that $\rho_q$ admits an envelope representation
\[
\rho_q(r) = \inf_{u>0}\left\{\frac{u}{2}\left(r-\frac{1-2q}{u}\right)^2 - \psi(u)\right\},
\]
and that the conditional mode of the auxiliary variable is
\[
\hat u(r)=\mathrm{sgn}(r)/r.
\]
In practice this yields per-observation weights $\omega_i=\hat u(r_i)$ and a shifted working response $z_i = y_i - (1-2q)/\omega_i$, so that the inner update becomes a weighted least-squares problem (plus the usual regularizer).
This directly connects to our main problem: once a difficult loss can be expressed through auxiliary variables as a weighted least-squares inner step, it fits the same computational template as WBB (random weights + fast optimization). Moreover, the quantile level $q$ plays the role of a tunable hyper-parameter; in the generator/transport-map view one can treat $(\omega,q,\lambda)$ as inputs and learn an amortized map $(\omega,q,\lambda)\mapsto \hat{\theta}$, enabling rapid hyper-parameter training over many $q$ (and $\lambda$) values without repeatedly solving the full optimization from scratch.

\subsection{Data augmentation and ECME for joint updates}

\citet{polson2011data} show how data augmentation ideas can be used for support vector machines and how ECME-style updates can jointly estimate parameters and regularization hyper-parameters. In their setup (with parameter of interest $\beta$ and regularization parameter $\nu$) one objective form is
\[
\sum_{i=1}^{n}(1-y_ix_i^T\beta)^+ + \nu^{-\alpha}\sum_{j=1}^{k} |\beta_j/\sigma_j|^{\alpha},
\]
with an inverse-gamma prior on $\nu^{-\alpha}$,
\[
p(\nu^{-\alpha})\propto \left(\dfrac{1}{\nu^{\alpha}}\right)^{a-1}\exp\left(-\dfrac{b}{\nu^{\alpha}}\right),
\]
leading to the update
\[
(\nu^{-\alpha})^{(g+1)} = \dfrac{b+\sum_{j=1}^{k}|\beta_j^{(g+1)}/\sigma_j|^{\alpha}}{a-1+k/\alpha}.
\]

\subsection{Generative multi-purpose sampler (GMS) for weighted M-estimation}

\citet{shin2024generative} study a generative multi-purpose sampler for weighted M-estimation, motivated by data perturbation schemes related to bootstrap and cross-validation, and provide implementations (including an R package and PyTorch code). A generic weighted objective is
\[
\hat{\theta}(\omega,\lambda)
= \arg\min_{\theta}\left\{
\frac{1}{n}\sum_{i=1}^n \omega_i\,\ell(\theta;y_i) + \lambda \phi(\theta)
\right\},
\]
which aligns naturally with the generator viewpoint above.

\subsection{Double descent and GCV for ridge tuning}

\citet{golub1979generalized} propose generalized cross-validation (GCV) as a method for choosing a ridge parameter. This classical lens is also closely related to modern discussions of double descent phenomena in interpolation regimes \citep{belkin2019reconciling,polson2025bayesian}. For ridge regression, the hat matrix can be written as
\[
A(\lambda) = X (X^T X + n\lambda I)^{-1} X^T.
\]
One can estimate $\lambda$ by minimizing a GCV criterion $V(\lambda)$. Even when $p>n$, the computation can be arranged to avoid inverting a $p\times p$ matrix directly (using identities that work with $n\times n$ objects).

\section{Empirical illustrations}

The goal of this section is to provide controlled, fully reproducible illustrations of the proposed template and its computational tradeoffs. These experiments are not intended as a final large-scale benchmark; rather, they demonstrate the mechanics of tuning criteria and amortized evaluation in settings where the inner optimizers are well understood.

\subsection{Ridge tuning: GCV, $K$-fold CV, and amortized evaluation}

We simulate a ridge regression problem with correlated design and compare (i) GCV-selected $\lambda$ \citep{golub1979generalized}, (ii) standard $K$-fold CV, and (iii) an amortized ``generator proxy'' trained to map $\lambda$ to coefficients and thereby approximate the CV curve without repeated refits across $\lambda$. To avoid data leakage, the generator is trained separately for each fold using only that fold's training data. The generator proxy is simple (a linear map from $(1,\log\lambda, (\log\lambda)^2)$ to $\beta$) to isolate the amortization idea. 

\begin{figure}[t]
\centering
\includegraphics[width=0.85\linewidth]{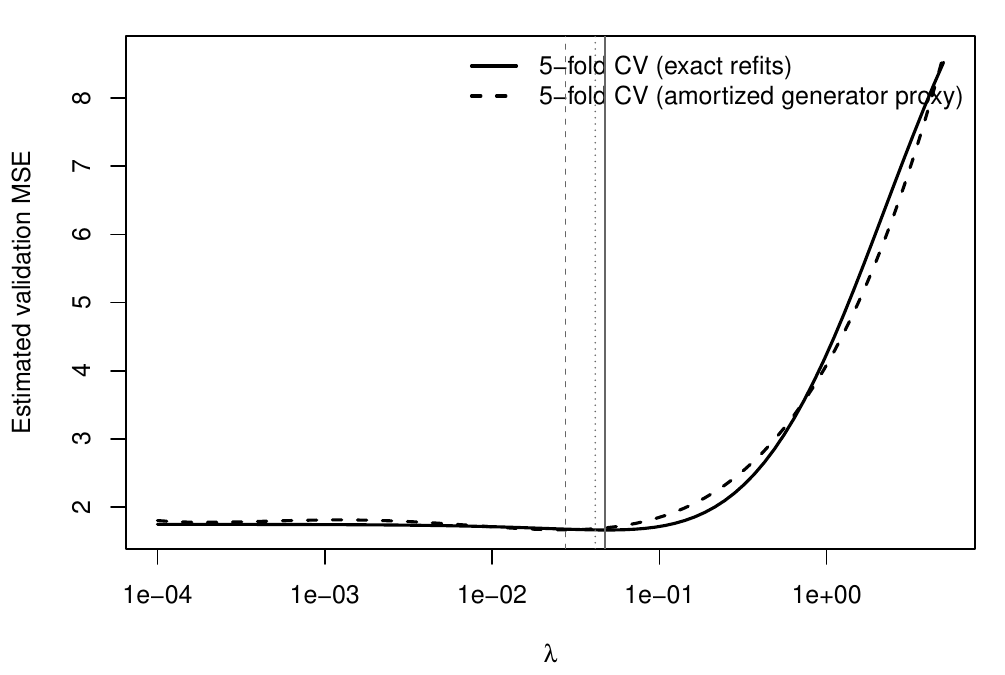}
\caption{Ridge tuning illustration. Solid line: 5-fold CV requiring repeated refits for each $\lambda$. Dashed line: a simple amortized proxy that approximates the CV curve by mapping $\lambda$ to coefficients via a learned map. Vertical lines indicate the selected $\lambda$ values.}
\label{fig:ridge-tuning}
\end{figure}

Figure~\ref{fig:ridge-tuning} shows the estimated CV curves; Table~\ref{tab:ridge-tuning} reports the selected $\lambda$ and test MSE. The amortized approach achieves a 4x speedup (0.03s vs 0.12s) while selecting a comparable regularization parameter. The slight degradation in performance is due to the limited capacity of the simple quadratic generator, illustrating the trade-off between amortization speed and approximation accuracy. The corresponding code is in \texttt{code/experiments/ridge\_tuning\_demo.R}.

\begin{table}[t]
\centering
\caption{Ridge tuning illustration (simulated data). We compare generalized cross-validation (GCV), $K$-fold cross-validation (CV), and a simple amortized generator proxy that approximates the CV curve without repeated refits across $\lambda$. Test MSE is computed on an independent hold-out test set.}
\label{tab:ridge-tuning}
\begin{tabular}{lcc}
\hline
Method & Selected $\lambda$ & Test MSE\\
\hline
GCV & 0.041 & 1.463 \\
5-fold CV & 0.047 & 1.450 \\
Amortized CV proxy (refit at selected $\lambda$) & 0.027 & 1.500 \\
\hline
\end{tabular}
\end{table}

\subsection{Deep Learning: Amortized tuning for MNIST}

To demonstrate the scalability of the generative approach, we apply the template to a deep learning task. We train a generator (hyper-network) to map the regularization strength $\lambda$ to the weights $\theta$ of a multi-layer perceptron (MLP) for MNIST classification. The generator is trained using the criterion-based objective (mode B) to minimize the weighted data loss plus $\lambda$-penalty across a range $\lambda \in [10^{-5}, 10^{-1}]$. 

\begin{figure}[t]
\centering
\includegraphics[width=0.85\linewidth]{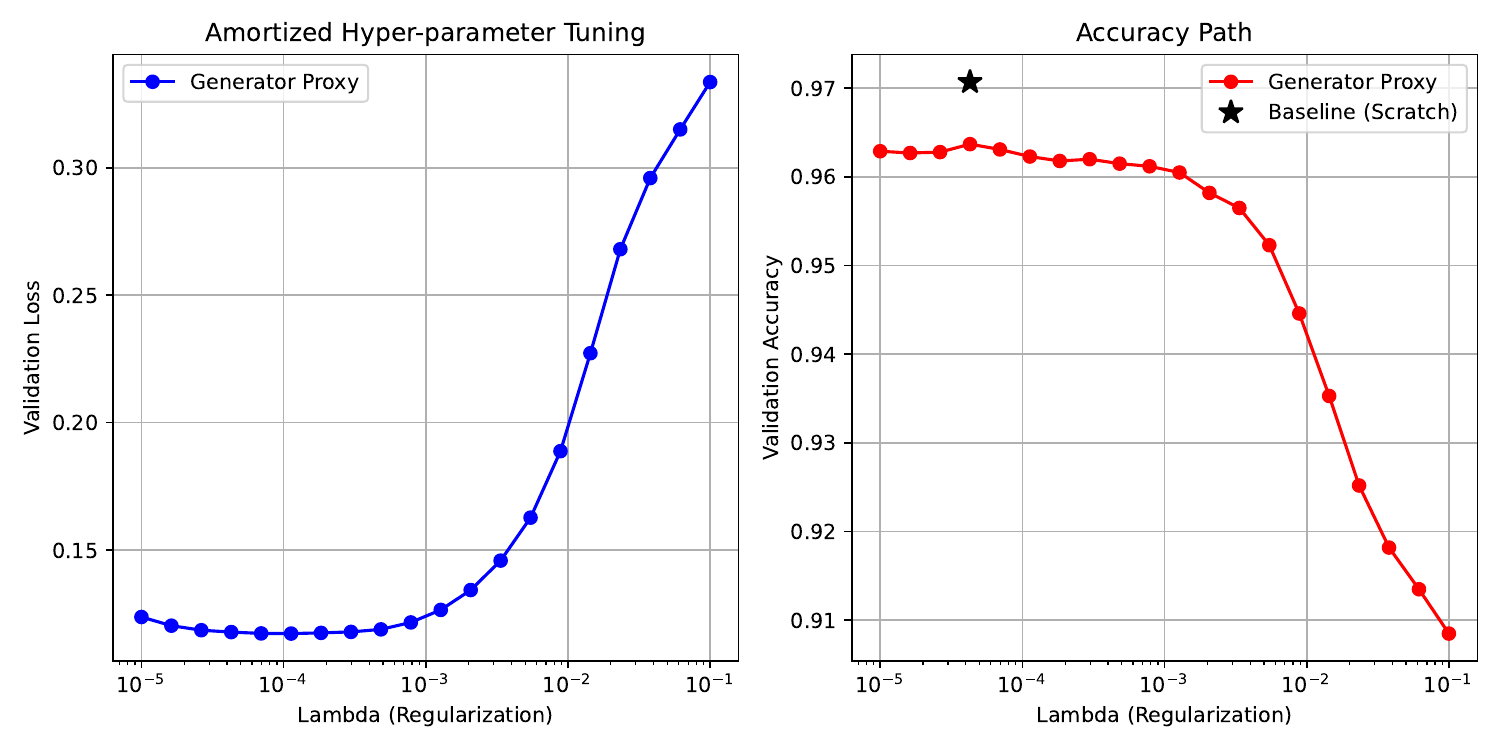}
\caption{MNIST tuning illustration using a generator network. The left panel shows the validation loss path and the right panel shows the validation accuracy path as functions of $\lambda$. The entire curve is produced by a single forward pass through the generator for each $\lambda$, avoiding expensive repeated training of the base model.}
\label{fig:mnist-tuning}
\end{figure}

Figure~\ref{fig:mnist-tuning} shows the validation loss and accuracy paths generated by the amortized map. The generator effectively captures the trade-off between regularization and fit, allowing one to "scroll" through hyper-parameter settings and identify the optimal $\lambda$ without re-training the MLP from scratch. Table~\ref{tab:mnist-tuning} reports representative performance metrics. 

Critically, the amortized approach offers massive computational savings for tuning. Evaluating the validation curve over 20 $\lambda$ values took approximately 8 seconds using the trained generator. In contrast, training a single baseline model from scratch at the optimal $\lambda$ took 13 seconds. Constructing the full tuning curve (20 points) without amortization would require roughly $20 \times 13 = 260$ seconds, implying a speedup of factor $>30$. The accuracy of the generator-produced model (96.2\%) is very close to the baseline model trained from scratch (97.1\%), validating that the amortization penalty is minimal. The implementation is available in our repository at \url{https://github.com/VadimSokolov/Generative-Bayesian-Hyperparameter-Tuning}.

\begin{table}[h]
\centering
\caption{Performance of the generator-produced models across different $\lambda$ values for MNIST.}
\label{tab:mnist-tuning}
\begin{tabular}{ccc}
\hline
$\lambda$ & Val Loss & Val Acc \\
\hline
1.0e-05 & 0.1238 & 0.9629 \\
1.1e-04 & 0.1173 & 0.9623 \\
1.3e-03 & 0.1265 & 0.9605 \\
1.4e-02 & 0.2273 & 0.9353 \\
1.0e-01 & 0.3336 & 0.9085 \\
\hline
Baseline ($\lambda=4.3e-05$) & - & 0.9707 \\
\hline
\end{tabular}

\end{table}

\section{Discussion}

This paper develops a unified view of hyper-parameter tuning as an outer criterion evaluated under an induced (possibly randomized) inner optimization map, and highlights amortization via generators/transport maps as a way to make repeated tuning and uncertainty summaries computationally feasible. By connecting GCV, Weighted Bayesian Bootstrap (WBB), and GMS-style criterion-based training, we provide a coherent framework for scalable hyper-parameter learning.

The central benefit of the proposed approach is the amortization of the optimization cost. As demonstrated in our MNIST experiment, constructing a tuning curve with a trained generator is orders of magnitude faster than retraining the base model from scratch for each candidate hyper-parameter. This speedup is particularly valuable for exploratory analysis, allowing researchers to interactively visualize the effect of regularization on performance metrics without waiting for new training runs. It also naturally extends to high-dimensional tuning, where the generator can accept vectors of hyper-parameters (e.g., layer-specific penalties, dropout rates) to explore spaces where grid search is infeasible. Furthermore, the randomized nature of the WBB objective means that a single generator can produce both point estimates (by fixing weights) and posterior samples (by resampling weights), essentially for free at test time.

Our ridge regression experiment illustrated a key trade-off: a simple generator (linear or quadratic map) provides massive speedups but may fail to capture the complex, non-linear path of the true optimizer, potentially leading to suboptimal hyper-parameter selection. In deep learning, however, over-parameterized networks often exhibit smooth dependence on hyper-parameters, which a neural generator (hyper-network) can approximate well, as seen in the MNIST results. Ensuring the generator has sufficient capacity relative to the complexity of the hyper-parameter-to-parameter map is crucial.

Several avenues for future work remain. First, adaptive sampling strategies for $(\lambda, \omega)$ during generator training could focus computational effort on the most relevant regions of hyper-parameter space. Second, architectural choices for the generator (e.g., rank-1 modulations, LoRA-style adapters) could further reduce the overhead of training the transport map. Finally, extending this framework to non-differentiable hyper-parameters (e.g., architecture search) remains a challenging but exciting frontier.

\bibliography{HyperTuning} 
\end{document}